\documentclass[journal]{vgtc}         

\usepackage{mathptmx}
\usepackage{graphicx}
\usepackage{times}
\usepackage{enumitem}
\setitemize{noitemsep,topsep=3pt,parsep=3pt,partopsep=3pt}

\usepackage[bookmarks,linkcolor=black]{hyperref} 
\hypersetup{
  pdfauthor = {},
  pdftitle = {},
  pdfsubject = {},
  pdfkeywords = {},
  colorlinks=true,
  linkcolor= black,
  citecolor= black,
  pageanchor=true,
  urlcolor = black,
  plainpages = false,
  linktocpage
}

\onlineid{101}

\vgtccategory{Research}

\vgtcinsertpkg

\title{Do Convolutional Neural Networks Learn Class Hierarchy?}

\author{Bilal Alsallakh, Amin Jourabloo, Mao Ye, Xiaoming Liu, Liu Ren}
\authorfooter{
\item
 Bilal Alsallakh, Mao Ye, and Liu Ren are with Bosch Research North America, Palo Alto, CA. E-mail: bilal.alsallakh@us.bosch.com, mao.ye2@us.bosch.com, liu.ren@us.bosch.com
\item
 Amin Jourabloo and Xiaoming Liu are with Michigan State University. E-mail: jourablo@msu.edu and liuxm@cse.msu.edu.
}

\shortauthortitle{Alsallakh et al. - Do Convolutional Neural Networks Learn Class Hierarchy?}

\abstract{

Convolutional Neural Networks (CNNs) currently achieve state-of-the-art accuracy in image classification.
With a growing number of classes, the accuracy usually drops as the possibilities of confusion increase.
Interestingly, the class confusion patterns follow a hierarchical structure over the classes.
We present visual-analytics methods to reveal and analyze this hierarchy of similar classes in relation with CNN-internal data.
We found that this hierarchy not only dictates the confusion patterns between the classes, it furthermore dictates the learning behavior of CNNs.
In particular, the early layers in these networks develop feature detectors that can separate high-level groups of classes quite well, even after a few training epochs.
In contrast, the latter layers require substantially more epochs to develop specialized feature detectors that can separate individual classes.
We demonstrate how these insights are key to significant improvement in accuracy by designing hierarchy-aware CNNs that accelerate model convergence and alleviate overfitting.
We further demonstrate how our methods help in identifying various quality issues in the training data.

}


\CCScatlist{ 
 \CCScat{K.6.1}{Management of Computing and Information Systems}%
{Project and People Management}{Life Cycle};
 \CCScat{K.7.m}{The Computing Profession}{Miscellaneous}{Ethics}
}

  \teaser{
 \centering
 \includegraphics[width=15.8cm]{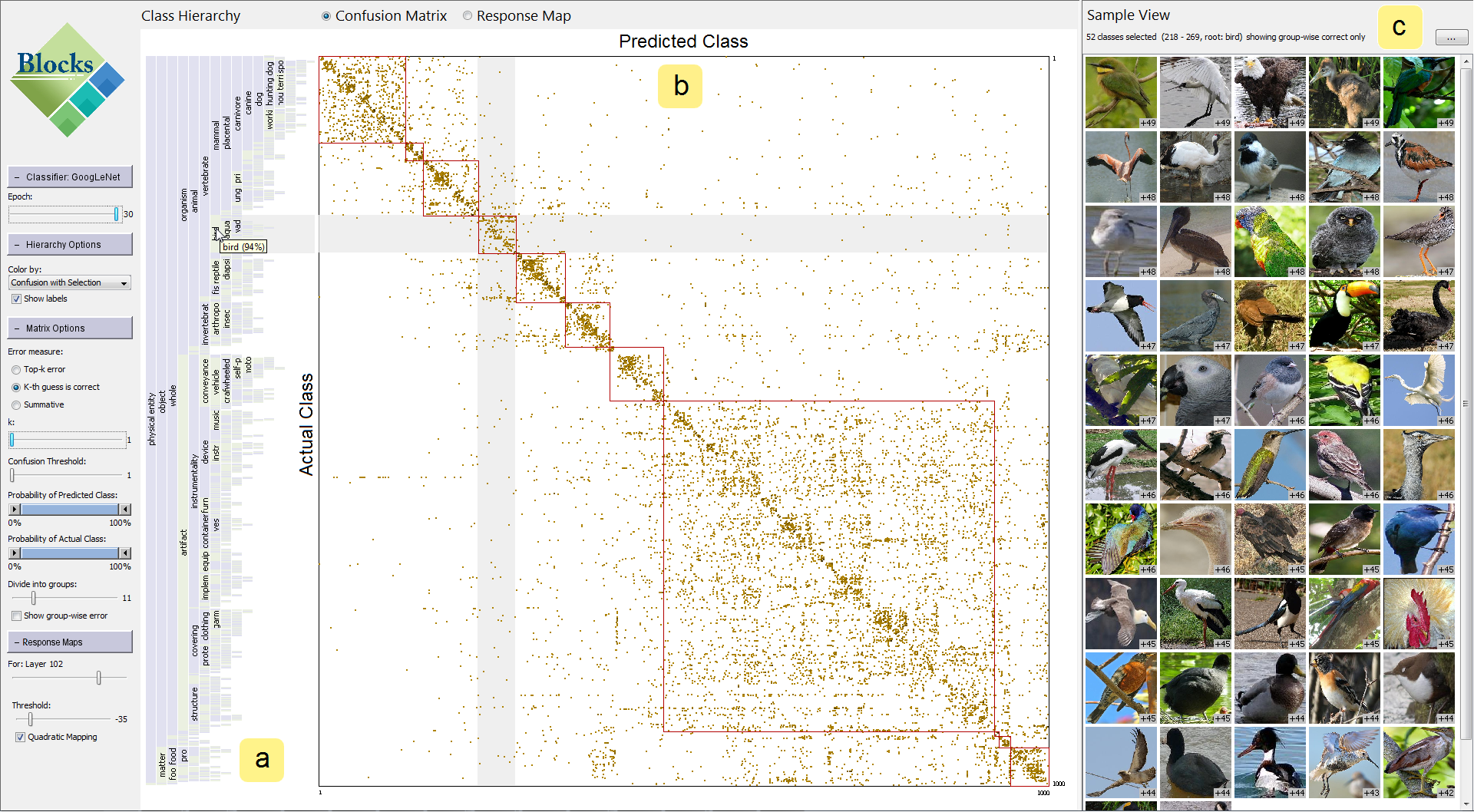}
\label{fig:teaser}
  \caption{
	The user interface of our system, showing classification results of the ImageNet ILSVRC dataset~\cite{russakovsky2015imagenet} using GoogLeNet~\cite{szegedy2015going}.
	(a) The class hierarchy with all classes under \emph{bird} group selected.
	(b) The confusion matrix showing misclassified samples only.
	The bands indicate the selected classes in both dimensions.
	(c)	The sample viewer shows selected samples grouped by actual class.
	}
  }

\begin{document}


\firstsection{Introduction}

\maketitle

\label{sec:intro}

Object recognition is a fundamental problem in computer vision that involves classifying an image into a pre-defined number of classes. 
Convolutional Neural Networks (CNNs) have achieved state-of-the-art results on this problem, thanks to the availability of large and labeled datasets and of powerful computation infrastructure  \cite{lecun2015deep}.
CNNs automatically extract discriminative classification features from the training images and use them in combination to recognize complex objects.
This enables CNNs to significantly outperform traditional computer vision approaches on large-scale datasets such as ImageNet \cite{deng2009imagenet}, as the latter usually rely on heuristic features \cite{dalal2005histograms, lowe1999object}.

To make CNNs applicable to critical domains, it is important to evaluate the reliability of the features they learn and to understand possible reasons behind classification errors \cite{ribeiro2016should}.
A number of powerful techniques have been proposed to visualize these features in the image space.
These visualizations demonstrate the power of these features and support the analogy between CNNs and natural vision systems.
However, little focus has been given to visualize the classification error itself and to refine CNNs accordingly.

We repeatedly observed that classification error follows a hierarchical grouping pattern over the classes.
We present a visual-analytics system, called \emph{Blocks}, to investigate this class hierarchy and to analyze its impact on class confusion patterns and features developed at each layer in the CNN.
\emph{Blocks} integrates all three facets of classification data when inspecting CNNs: input samples, internal representations, and classification results.
It enables scalable inspection of these facets, at the scale of ImageNet, in order to:
\begin{itemize}
	\item Identify various sources of classification error (\textbf{T1}).
	\item Exploit the hierarchical structure of the classes to improve the CNN architecture, training process, and accuracy (\textbf{T2}).
	\item Analyze the CNN's sensitivity to data variation and curate a balanced training data that improves its robustness (\textbf{T3}).
\end{itemize}
These tasks involve the high-level goals of visualizing machine-learning data as characterized by Liu et al.~\cite{liu2017towards2}: \emph{understand}, \emph{diagnose}, and \emph{improve}.
Section \ref{sec:apps} illustrates how \emph{Blocks} enables these tasks and reports quantitative results of how involving the class hierarchy reduces the top-5 error of a reference CNN by more than one third.





\section{Motivation and Background}

\label{sec:motivation}

The yearly ImageNet Large Scale Visual Recognition Competition (ILSVRC) challenges participants to classify images into one thousand object categories chosen randomly from ImageNet \cite{russakovsky2015imagenet}.
In 2012, Krizhevsky et al.~\cite{krizhevsky2012imagenet} trained a CNN classifier which won the competition by a large margin.
This led to a paradigm shift in computer vision, with extensive research to understand how CNNs work. 

We examined classification error of publically-available CNNs, pre-trained on the ILSVRC 2012 training set.
For this purpose we generated confusion matrices which show how often a pair of classes are confused for each other when classifying the corresponding validation set. 
By re-ordering the rows and columns of these matrices by similarity, we consistently found two major blocks along the diagonal which contain more than $98\%$ of misclassifications.
One block corresponds to natural objects such as plants and animals, while the other block represents artifacts such as vehicles and devices.
This means that CNNs rarely confuse natural objects for artifacts or vice versa.
By reordering each block individually, we found that it in turn contains sub-blocks that capture the majority of confusions.
This sparked our interest to  investigate how these structures can be exploited to improve classification accuracy of CNNs.

\subsection{ImageNet and the ILSVRC 2012 Dataset}
\label{sec:dataset}
Curated in 2009, ImageNet is the largest publically available labeled image dataset, encompassing more than 14 million images that belong to more than $20,000$ object categories \cite{deng2009imagenet}.
The object categories are nouns in the WordNet database of the English language \cite{miller1995wordnet} .

A fundamental property of WordNet is its hierarchical organization of concepts, e.g. \emph{birds} are \emph{vertebrates}, \emph{vertebrates} are \emph{organisms}, and so on.
The $1000$ classes of the ILSVRC 2012 dataset are leaf nodes in this hierarchy that are randomly selected according to certain criteria that aim to reduce ambiguities.
The dataset contains about 1.2 million images in the training set in addition to $50,000$ images in the validation set.
By ordering these classes according to the WordNet hierarchy, we found the same class grouping structure we observed in the confusion matrices (Fig.~1).
After examining the CNN classifiers, we found that they surprisingly did not make any use of the class hierarchy information in the training phase.
Deng et al.~\cite{deng2010does} made a similar observation after comparing a number of classifiers on ImageNet, concluding that visual object categories are naturally hierarchical.
In this work we examine how this hierarchical structure impacts CNNs.

\subsection{Convolutional Neural Networks (CNNs)}

CNNs are a special type of feed-forward neural networks that contain a number of \emph{convolutional} layers.
A convolutional layer consists of learnable filters that respond to certain features when convolved with a 2D input, producing a filtered 2D output.
The first convolutional layer is applied to the input image, whereas subsequent layers take the output of the respective preceding layer as input (Fig.~\ref{fig:hierarchicalcnn}).
Special layers are inserted between convolutional layers to reduce the dimensionality and to add necessary non-linearity \cite{lecun1998gradient}.

After training a CNN, the convolutional filters become feature detectors in the image.
Appropriate visualization techniques can reveal the features these filters respond to, as we explain next.

\subsection{State of the Art in Visualizing CNNs}

\label{sec:relatedwork}

Visualization has played a major role in understanding and optimizing CNNs.
A major focus has been made on visualizing the image features each filter learns to detect. 
Further techniques have addressed different aspects of the data involved in the CNN pipeline. %

\subsubsection{Feature Visualization}
\label{sec:featurevis}
Image-based visualizations are a natural way to inspect the feature detectors developed by a CNN.
Various techniques have been proposed for this purpose, 
based on four main approaches \cite{grun2016taxonomy, nguyen2016multifaceted, mahendran2016visualizing}:

 \begin{itemize}
 	\item \textbf{Input modification / occlusion}: 
	these techniques aim to reveal which regions in real images contribute most to a target response.
	This is done by occluding different regions of the input image individually and measuring the impact on the target using a forward pass \cite{zeiler2014visualizing, zhou2014object}.
	The result is usually a coarse 2D saliency map (also called activation map) which visualizes the importance of each region of the image to the target response. 
	

	\item \textbf{Deconvolution}: 
	these techniques also aim to find which parts in real images contribute most to a target response.
	In contrast to input modification, the response is traced \emph{backwards} to the input image by reversing the operations performed at each layer using various methods \cite{bach2015pixel, simonyan2014deep, springenberg2014striving, zeiler2014visualizing}.
	This produces a fine-grained saliency map of the input image at the pixel level, which in turn reveals the learned features and their structure (Fig.~\ref{fig:featurevis}).
	
	\item \textbf{Code inversion}:
	these techniques first apply the CNN to a real image and compute the collective response, called the code, of a particular layer.
	An image is then synthesized that would create a similar response at this layer	using various methods \cite{mahendran2015understanding, dosovitskiy2016inverting, mahendran2016visualizing}.
	Comparing both images reveals which features of the real image are retained at the selected layer.
	Caricaturization can further emphasize these features \cite{mahendran2016visualizing}.

	\item \textbf{Activation maximization}:
	these techniques, unlike previous ones, do not require a real image.
	Instead, they synthesize an artificial image that, if used as input, would maximize a target response.
	Early techniques often produced synthetic images that were hard to recognize \cite{simonyan2014deep, yosinski2015understanding}.
	Accounting for the multi-faceted nature of neurons \cite{ nguyen2016multifaceted, wei2015understanding} and imposing natural image priors \cite{mahendran2016visualizing, nguyen2016synthesizing} have significantly improved the interpretability. 
\end{itemize}

The above techniques were shown useful in diagnosing CNNs and in refining the architecture accordingly \cite{zeiler2014visualizing}.
However, they fall short of exposing high-level patterns in collective CNN responses computed for images of all classes.

\subsubsection{Projection-based Visualization}

\label{sec:scatterplots}

These techniques aim to provide overview of network-internal data by projecting them into a 2D space using various projection methods.

A typical use of projection is to assess class separability at different layers or at different iterations during training \cite{donahue2014decaf, zhang2016efficient, rauber2017visualizing}.
This helps in comparing classification difficulty of different datasets as well as identifying under-fitting and over-fitting models.
When the projected data correspond to the input images, icons of these images can be used instead of dots \cite{aubry2015understanding, nguyen2016multifaceted, oh2016deep}.
This helps in observing visual patterns in the samples, such as possible latent subclasses among the samples of one class (e.g. red and green peppers) ~\cite{aubry2015understanding, nguyen2016multifaceted}.
Scatter plots have also been used to reveal similarities between neurons \cite{chungrevacnn, rauber2017visualizing} and to compare learning trajectories of multiple networks~\cite{erhan2010does}.

\subsubsection{Network-based Visualization}

Many techniques emphasize the feed-forward structure in CNNs by showing neurons as nodes in successive layers connected by links, and mapping data facets on top of them.
This has been shown useful to inspect how the network classifies a selected or user-generated sample \cite{Harley2015, tzeng2005opening}.
\emph{{ReVACNN}}~\cite{chungrevacnn} enables inspecting how neuron activations develop during training.
\emph{Tensorflow Playground} \cite{smilkov2016directmanipulation} 
enables novice users to construct a network and interactively observe how it learns to separate between two classes in datasets of varying  difficulty.
\newline\emph{CNNVis}~\cite{liu2017towards} is a powerful system designed to diagnose deep CNNs.
It employs various clustering algorithms to group neurons in representative layers based on response similarity and to group connections between these layers accordingly.
A neuron cluster can be augmented with thumbnail images showing stimuli that activate these neurons most.
The authors demonstrate how \emph{CNNVis} exposes a variety of issues in network design such as redundancies in layers and neurons, as well as inappropriate learning parameters. 

Finally, several frameworks offer a visualization of network architecture \cite{saxena2016convolutional, yeager2016effective}.
This is useful to comprehend large networks and to compare multiple architectures.

\subsubsection{Training Data and Performance Visualization}

The majority of previous work focused on CNN-internal data as key to understand and optimize CNNs.
Besides appropriate architecture and learning parameters, the quality of training data is also essential to learning generalizable CNNs models.
Inspecting the quality of training data is nontrivial especially due to the large volume of data needed to train CNNs.
In an attempt to address this issue, NVIDIA released \emph{DIGITS}, a system that enables users to browse image datasets and inspect images of a certain class \cite{yeager2016effective}.
Users can apply image transformations such as cropping or resizing to match the CNN input size while preserving important parts of the image.
Additional plots such as line charts and confusion matrices allow inspecting the performance.
The system is limited to datasets encompassing a few dozens of classes, and does not link performance results with the input data.

Our work aims to fill the gap in available tools by offering an integrated exploration environment to analyze all three data facets involved in the CNN pipeline: input images, CNN-internal data, and classification results.
Offering this integration in a scalable way is key to an advanced analysis of large-scale CNNs and to close the analysis loop by guiding model refinements that improve the accuracy.
	

\section{\emph{Blocks}}
\label{sec:blocks}


Being the target of classification, the class information is the most salient information along the classification pipeline.
It is present both in the labeled input and in the output, and it largely determines the features learned by the CNN.
Classes have varying degrees of discriminability.
Some classes have unique features such as strawberries and zebras, while other classes might share similar features and are hence harder to distinguish from each other.
Hinton et al.~\cite{hinton2015distilling} noted that such similarity structures in the data are very valuable information that could potentially lead to improve classifiers.
Our work offers new means to analyze these structures and their impact on CNNs.

With a growing number of classes, the similarity structures between them become complex. 
As we mentioned in Section~\ref{sec:motivation}, a key observation about these structures is their hierarchical nature: classes within the same branch of the hierarchy are increasingly more similar to each other than to other classes.
We designed our visual analytics system around this idea.
In particular, we focus on revealing the hierarchical similarity structure among the classes and on analyzing how it impacts both the classification results and the image features the CNN learns to recognize.
We call our system \emph{Blocks} as it extensively relies on visual block patterns in identifying similarity groups. 


The main interface of \emph{Blocks} consists of four views that show different facets of the data:
the \emph{hierarchy viewer} (Fig.~1a), the \emph{confusion matrix} (Fig.~1b), the \emph{response map} (Fig.~\ref{fig:featuremap}c), and the \emph{sample viewer} (Fig.~1c).
The first three views show information aggregated at the class level and use a unified class order, dictated by the class hierarchy.
The sample viewer shows image samples according to user selections in the other views.
Each view contributes in certain ways to the high-level analysis tasks \textbf{T1}-\textbf{T3} listed in Section~1.
At a time, the user can display either the confusion matrix or the response map as the active view.
The hierarchy viewer is displayed to the left of the active view and indicates the class ordering along the vertical axis.

The class hierarchy can be either pre-defined or constructed interactively with help of the confusion matrix (Section~\ref{sec:ordering}).
The next sections describe the above-mentioned views, illustrated on the ILSVRC 2012 dataset, classified using GoogLeNet \cite{szegedy2015going}.
This dataset has a pre-defined class hierarchy, as explained in Section~\ref{sec:dataset}.

\subsection{Class Hierarchy Viewer}
\label{sec:hierarchy}

\emph{Blocks} shows the class hierarchy using a horizontal icicle plot \cite{kruskal1983icicle} along the vertical dimension (Fig.~1).
Each rectangle in this plot represents a group of classes.
The rectangle color can encode information about this group such as a group-level performance metric (Fig.~\ref{fig:matevolution}).
These metrics are computed by considering the groups to be the classification target.
A sample is correctly classified with respect to a group if both its actual and predicted classes are in the group.
This enables defining the following metrics:
 \begin{itemize}
	\item \textbf{Group-level precision}: this measures how many of the samples classified in a group actually belong to the group.
 	\item \textbf{Group-level recall}: this measures how many of the samples that actually belong to a group are classified into the group.
	\item \textbf{Group-level F-measure}: this can be defined based on group-level precision and recall as follows:
		\begin{equation}
			F_1(g) = 2 \cdot \frac{ \mbox{Precision} (g) \cdot \mbox{Recall}(g)}{\mbox{Precision}(g) + \mbox{Recall}(g)} 
		\label{eq:F-measure}
		\end{equation}
 \end{itemize}
As we show in Section~\ref{sec:apps}, inspecting group-level performance under different conditions reveals the impact of the hierarchical structure on CNN performance (\textbf{T2}) and its sensitivity to data variation (\textbf{T3}).

The child nodes of a parent node in the hierarchy can be sorted by a user-selected criterion, such as size or performance metrics.
Nodes that have only one child are contracted to compress the hierarchy and reduce the visual complexity.
Hovering the mouse over a rectangle shows information about the respective group including its label and performance metrics.
Clicking on a rectangle selects the corresponding classes and updates the other views to focus on these classes.
This enables inspecting their samples and analyzing their confusion patterns and CNN-internal responses.


\subsection{Confusion Matrix}


Confusion matrices have been utilized in the machine learning community for various purposes such as detailed comparison of performance and identifying frequent confusion between certain classes.
We argue that these matrices can reveal further information about error structure (\textbf{T1} and  \textbf{T2}) and classifier behavior (\textbf{T2}) when equipped with appropriate ordering, visual encoding, and user interactions.


\subsubsection{Class ordering - constructing the class hierarchy} 
\label{sec:ordering}
A confusion matrix is re-orderable \cite{bertin1983semiology}, as long as the same class order is used along the rows and columns.
This ensures that the correct classifications are encoded along the matrix diagonal.
The desired ordering should reveal similarity groups among the classes.
This corresponds to a \emph{block pattern} in the matrix \cite{behrisch2016matrix}: the majority of confusion takes places within a number of blocks along the diagonal, each of which corresponds to a similarity group of classes.

In case a pre-defined class hierarchy is available, \emph{Blocks} displays it in the hierarchy viewer and orders the matrix accordingly.
If such a hierarchy is unavailable or fails to reveal a block pattern, the user can explore if such pattern exists by interactively applying a seriation algorithm.
Behrisch et al.~\cite{behrisch2016matrix} surveyed various seriation algorithms that can reveal block patterns in matrices.
\emph{Blocks} offers both fast algorithms \cite{hubert1974some, makinen2005barycenter} and exhaustive ones such as spectral clustering \cite{guattery1998quality}.

The hierarchy can be refined recursively, as proposed by Griffin and Perona \cite{griffin2008learning}: the user selects a high-level block and applies the algorithm on this part.
At each step, the matrix is updated to allow inspecting the plausibility of the computed sub-blocks and to guide algorithmic choices.
If plausible, the hierarchy viewer is updated to reflect the constructed hierarchical structure.


After the class hierarchy and the corresponding block patterns are established, it is possible to distinguish between non-diagonal matrix cells based on their location in the matrix: Cells that are within a dense block represent confusions between highly-similar classes.
Cells that do not belong to a block represent unexpected confusions between classes that seem to be less related, and are hence especially interesting to explore further (Section~\ref{sec:data_quality}).
We call these cells \emph{block outliers}.



\subsubsection{Visual encoding} 
Besides an appropriate class ordering, the visual encoding of the cell values plays a major role in revealing block patterns and their outliers.
In machine-learning literature, confusion matrices are often generated using the default \emph{Jet} color map in MATLAB \cite{escalera2014chalearn, joshi2012scalable, murthy2016deep}.
Instead, we use a sequential color scale which maps the value $1$ to a light shade and the largest value to a dark shade.
Cells with value $0$ remain white, which facilitates identifying and selecting non-zero cells that represent actual confusions (Fig.~1b and Fig.~\ref{fig:matrix_filtering}).

\paragraph{\textbf{Focusing on misclassification}}
By default, we exclude the matrix diagonal from the visual mapping since correct classifications usually account for the majority of the value sum in the matrix.
This eliminates an, otherwise, salient diagonal which interferes with fine-grained block patterns.
The per-class accuracy can be displayed more appropriately on top of the class hierarchy or in the sample viewer.

\paragraph{\textbf{Non-linear mapping}}
Even among off-diagonal cells, there is typically a large variation in values.
While the majority of non-zero cells typically have small values, 
a very small number of cells might have large values and indicate classes that are very frequently confused for each other.
To alleviate such variation, the user can select a logarithmic mapping of values to color, which helps emphasize less frequent confusions that form the block patterns.
Interactive filtering allows identifying cells that represent frequent class confusions.

\paragraph{\textbf{Visual boosting}}
Even though standard displays offer sufficient space to map a $1000 \times 1000$ matrix to pixels without overlaps, assigning one pixel to a cell makes it barely visible, which might leave block outliers unnoticed.
The user can select to emphasize non-zero cells by enabling a halo effect \cite{oelke2011visual}, which extends 1-pixel cells into $3 \times 3$ pixels and assigns $30\%$ opacity to the peripheral halo area.
This effect not only emphasizes block outliers, it further improves the perception of blocks and sub-blocks within them.
The halos are visual artifacts that might add shade to, otherwise, empty cells.
Individual confusions can hence be examined more precisely using interaction.

 


\subsubsection{Interaction} 

\emph{Blocks} enables various interactions with the confusion matrix. 
As we illustrate in the supplementary video, these interactions are essential to identify various sources of classification errors (\textbf{T1}), especially those related to data quality issues (Section~\ref{sec:data_quality}).

\paragraph{\textbf{Selection}} 
There are two ways to select samples in the matrix:
\begin{itemize}
	\item Drawing a box around certain cells.
		This updates the sample viewer to show the corresponding samples. 
	\item Clicking on a group in the class hierarchy.
		This highlights false positives (FPs) and false negatives (FNs) with respect to the group classes by means of vertical and horizontal bands (Fig.~1).
		The intersection of these bands are confusions between classes that belong to the selected group and hence represent group-level true positives (TPs).
		The difference of these bands corresponds to group-level FPs and FNs respectively.
		The sample viewer is updated to show the highlighted samples, and allows exploring the group-level TPs, FPs, and FNs individually.
\end{itemize}

\begin{figure}[!ht]
 \centering
 \includegraphics[width=\linewidth]{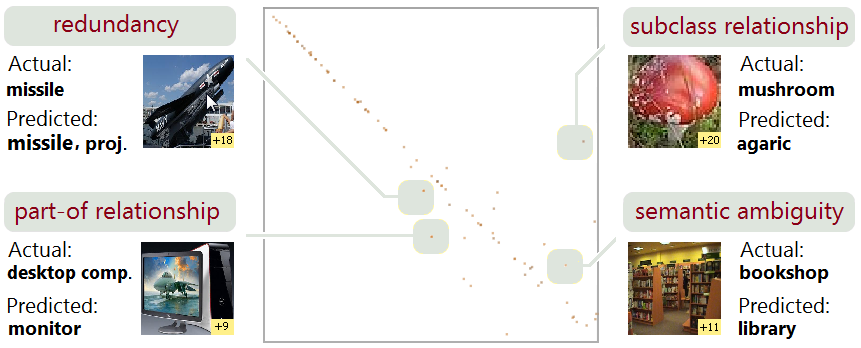}
 \caption {
	Filtering out diagonal cells and cells whose values are $<10$ to retain repetitive confusions.
	Near-diagonal cells correspond to highly similar classes while off-diagonal cells often indicate data quality issues.
 }
 \label{fig:matrix_filtering}
\end{figure}

\paragraph{\textbf{Filtering}} 
The mis-classified samples encoded in the matrix cells can be filtered according to multiple criteria.
The matrix is updated to show confusion patterns among the filtered samples. 
\begin{itemize}
	\item Filtering by cell value:
		This retains cells representing repetitive class confusions above a selected threshold (Fig.~\ref{fig:matrix_filtering}).
		These confusions often indicate overlapping class semantics (Section~\ref{sec:data_quality}).
	\item Filtering by top-k results:
	This filters out samples whose correct labels are among the top-k guesses computed by the classifier.
	The remaining samples represent the classifier's top-k error, a commonly-used performance measure that relaxes the requirement of correct classification by accepting multiple guesses.
	\item Filtering by classification probability:
		This retains samples for which the classifier predictions were computed with probability in a certain range.
		It is possible to further specify a range for the probability computed for the \emph{actual} class.
\end{itemize}

\paragraph{\textbf{Grouping}} 
\emph{Blocks} enables emphasizing the block pattern in the matrix by drawing boxes around major blocks (Fig.~1).
The user specifies the number of blocks $b$, which are then determined by a partitioning algorithm.
The algorithm selects a partitioning which maximizes the density of its blocks.
The boxes are retained during filtering, which helps keeping track of block memberships.
It is possible to divide the matrix into $b \times b$ clickable regions based on the blocks, which eases the identification and selection of block outliers.



\begin{figure*}[]
 \centering
 \includegraphics[width=\textwidth]{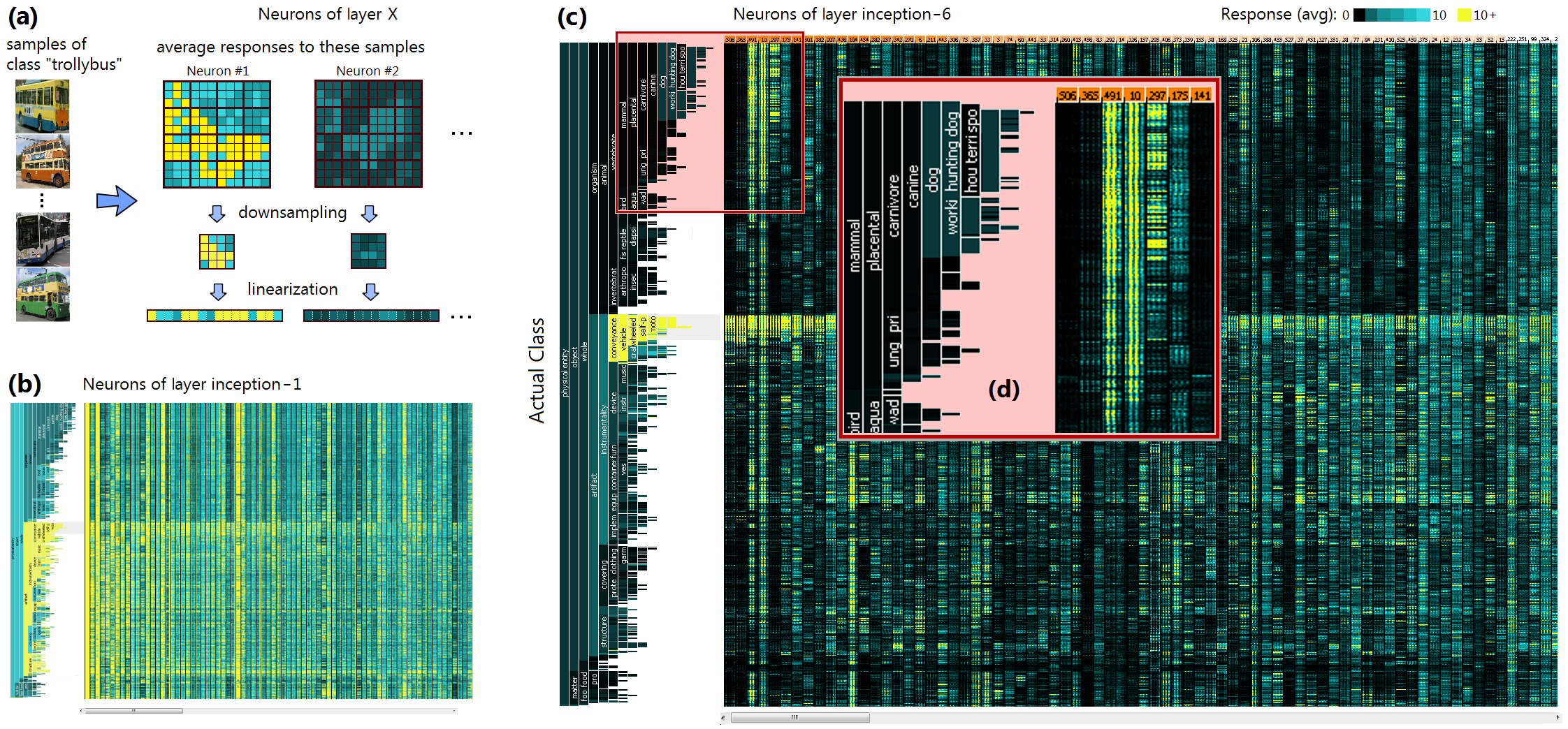}
 \caption {
	The Response Map:
	(a) Illustrating how the row that corresponds to class \emph{trollybus} is computed. Each column represents the average responses of a neuron in the selected layer.	
	(b, c) The response maps of layers \emph{inception}-1 and \emph{inception}-6  in GoogLeNet~\cite{szegedy2015going}.
	The rows represent the classes and are ordered by the class hierarchy depicted to the left of each map.
	The \emph{wheeled vehicle} group is selected, and the neurons are sorted by their relevance to it (Eq.~\ref{eq:neuronRelevance}).
	The most relevant neurons in layer inception-6 can separate the classes in this group from other classes, while inception-1 can only separate higher-level groups.
	 (d) Pose-based detectors  of \emph{vehicles} have high responses among \emph{mammals} as well.
 }
 \label{fig:featuremap}
\end{figure*}

\subsection{Response Map}
\label{sec:responsemap}

This view provides overview of the CNN responses at a selected layer to all samples in the dataset.
The aim is to identify whether classes in the same group activate a similar set of features, and which combination of features characterize a class or a group of classes.
This enables understanding how well different layers in the network can discriminate between groups in different levels of the class hierarchy (\textbf{T2}) and how sensitive to data variation the developed features are (\textbf{T3}).

As illustrated in Fig.~\ref{fig:featuremap}a, the neuron responses are averaged per class, over all of its samples.
This aims to reveal variations in these responses across \emph{classes} and \emph{neurons}, not across samples. 
This further enables a compact visual encoding of responses computed from a large number of samples. 
Responses to individual samples of a particular class can be explored on demand in an auxiliary view (Section~\ref{sec:sample_level_responsemap}).

\subsubsection{Visual encoding}

We use a heatmap to encode per-class average response of each neuron in the selected layer (Fig.~\ref{fig:featuremap}c).
The rows of the heatmap represent the classes and are ordered according to the class hierarchy.
The columns represent the neurons, and their order is updated according to user selection.
A neuron can have multiple output channels as in the case of filters in convolutional layers and the associated pooling units and rectified linear units (ReLUs).
\emph{Blocks} visualizes these channels as vertical 1-pixel-wide lines within the neuron's column.
This is done by linearizing these channels as illustrated in Fig.~\ref{fig:featuremap}a.
As a result, the 2-dimensional structure of the neuron's output is lost, in favor of emphasizing how its responses vary across multiple classes, which we denote as the \emph{response profile} of the neuron.

Cell color represents the average response of a neuron's channel among samples of a certain class.
The user can specify a threshold $T$ on this response.
Values smaller than $T$ are mapped linearly to a color scale from black to light blue.
Values equal to or larger than $T$ are shown in yellow.
This aims to emphasize cells representing high responses, in context of the other cells.
Adjusting the threshold allows identifying neurons that respond specifically to certain classes and exploring subtle differences between different response profiles.

In some CNNs, the convolutional filters can be as large as $64 \times 64$, especially in early layers.
To gain overview of multiple filters of this size in one view, \emph{Blocks} allows downsampling their output e.g. to $8 \times 8$.
Fig.~\ref{fig:featuremap}a  illustrates how the responses of a $12 \times 12$ filter are downsampled to $4 \times 4$ channels which fit in a 16-pixel-wide column.
This allows comparing multiple response profiles side by side.
Furthermore, this consolidates major variations between these profiles that  would be, otherwise, scattered across numerous channels.



\subsubsection{Exploring group-level features}
\label{sec:grouplevelfeatures}

The unified class ordering in \emph{Blocks} enables analyzing the relation between the response profiles of the neurons and the class hierarchy.
We observe that certain profiles show high responses mainly for samples within a particular group of classes in the class hierarchy.
This means that the corresponding neurons learned shared features among these classes such as shape, pose, or background.
As we illustrate in the supplementary video, interaction is key to identify neurons that respond to a particular group in the class hierarchy.
In Fig.~\ref{fig:featuremap}b-c, the columns are reordered according to the ability of the corresponding neurons to distinguish  \emph{wheeled vehicles} from the other classes.
For this purpose we compute a relevance measure $R_G(N)$ for each neuron $N$, based on its responses to group samples $G$ and to non-group samples $\overline{G}$:
\begin{equation}
	R_G(N) = \frac{Q_{1/4}(\{f_N(x) : x \in G\})}{Q_{3/4}(\{f_N(x) : x \in \overline{G}\})} 
\label{eq:neuronRelevance}
\end{equation}
where $f_N(x)$ is the collective response of the neuron to a sample $x$, computed as the sum of all of its output channels, and $Q_{i/q}$ is the $i$-th $q$-Quantile.
This measure mimics statistical significance tests and takes a high value when the response is consistently high among the group classes and consistently low among non-group classes.
The column headers can communicate the computed values via color.
Visual inspection enables identifying if a neuron responds to a sub-group or super-group of the selected group, or possibly to other groups as well.
For example, no neuron in the early layer \emph{inception}-1 can capture the selected group specifically (Fig.~\ref{fig:featuremap}b), unlike 
the advanced layer \emph{inception}-6 (Fig.~\ref{fig:featuremap}c).
Furthermore, certain neurons that respond to \emph{wheeled vehicles} respond highly to \emph{mammals} as well (Fig.~\ref{fig:featuremap}d).
These neurons detect pose features that are shared between both groups of classes.

\begin{figure*}[!ht]
 \centering
 \includegraphics[width=\textwidth]{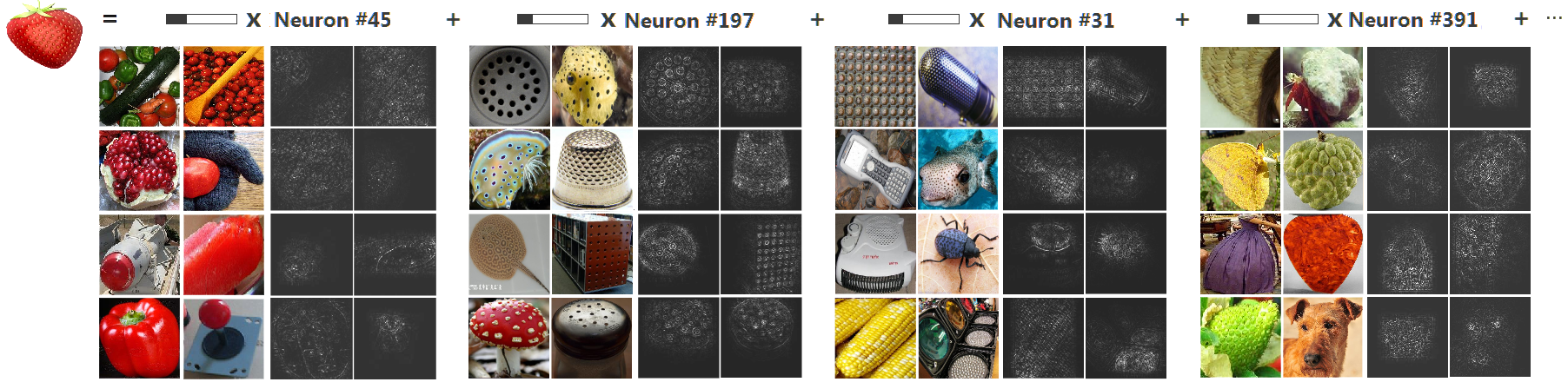}
 \caption {
		Feature detectors at layer \emph{inception-6} in GoogLeNet that show high response to samples of class \emph{strawberry}.
		We depict the top-9 images in ILSVRC validation set that activate each detector most, along with the corresponding saliency maps (computed using \emph{FeatureVis}~\cite{grun2016taxonomy}).
 }
 \label{fig:featurevis}
\end{figure*}

We found that group-level features are often based on shape, pose, and background.
For example, within natural objects, a combination of shape and pose features can distinguish high-level groups such as \emph{birds}, \emph{mammals}, and \emph{insects}.
Background features are involved in certain groups such as \emph{fishes} and \emph{geological formations}.
On the other hand, color features as well as certain texture features are often shared across various classes that do not fall in a specific group.
To facilitate analyzing such cases, the groups in the hierarchy viewer can be colored by the average response of a selected neuron (Fig.~\ref{fig:featuremap}b-c).

For some groups, such as \emph{devices}, no neuron exhibits significant difference in responses between group classes and non-group classes.
Such lack of group-level features indicates a high variation among the group classes that develop class-specific features instead.

\subsubsection{Exploring neuron properties}

\label{sec:features}

 Besides the response profiles, \emph{Blocks} provides additional information about a neuron either in summary or in detail forms.
 The header of the response map can communicate summary information about the neurons by means of color (Fig.~\ref{fig:featuremap}c-d).
 Examples for this are average activation within the samples of a selected class, relevance to a selected group, or sensitivity to an image transformation (Section~\ref{sec:sensitivity}).


Clicking on a profile header updates the sample viewer to show samples that highly activate the corresponding neuron.
This aims to help users find out common image features across these samples in order to identify the image features to which the neuron responds.
The sample viewer provides several possibilities to explore the samples along with saliency maps of their image features. 

Understanding the image features a neuron responds to is important to understand how each class is being detected by the CNNs and why certain samples of it are misclassified.
Typically, the network characterizes each class by a set of features that are detected by different neurons.
Fig.~\ref{fig:featurevis} illustrates image features that characterize the class \emph{strawberry} at an intermediate layer in GoogLeNet.
These features correspond to the four most relevant neurons to this class in this layer.
The first neuron detects red objects, the second and third neurons detect dotted objects and objects with bumps, and the fourth neuron detects natural objects having isosceles triangular shapes.
This means that strawberries are detected based on color, texture, and shape, in the respective order of importance.
We found that images of unripe strawberries and strawberry images in grayscale do not activate the first neuron and are therefore often misclassified (\textbf{T1}).
On the other hand, classes whose samples have varying colors such as vehicles do not rely on color.
Such findings are useful to curate training data that are representative of the target classes (\textbf{T3}) as we show in Section~\ref{sec:sensitivity}.




\subsubsection{Sample-level responses and latent subclasses}
\label{sec:sample_level_responsemap}

\begin{figure}[!b]
 \centering
 \includegraphics[width=\linewidth]{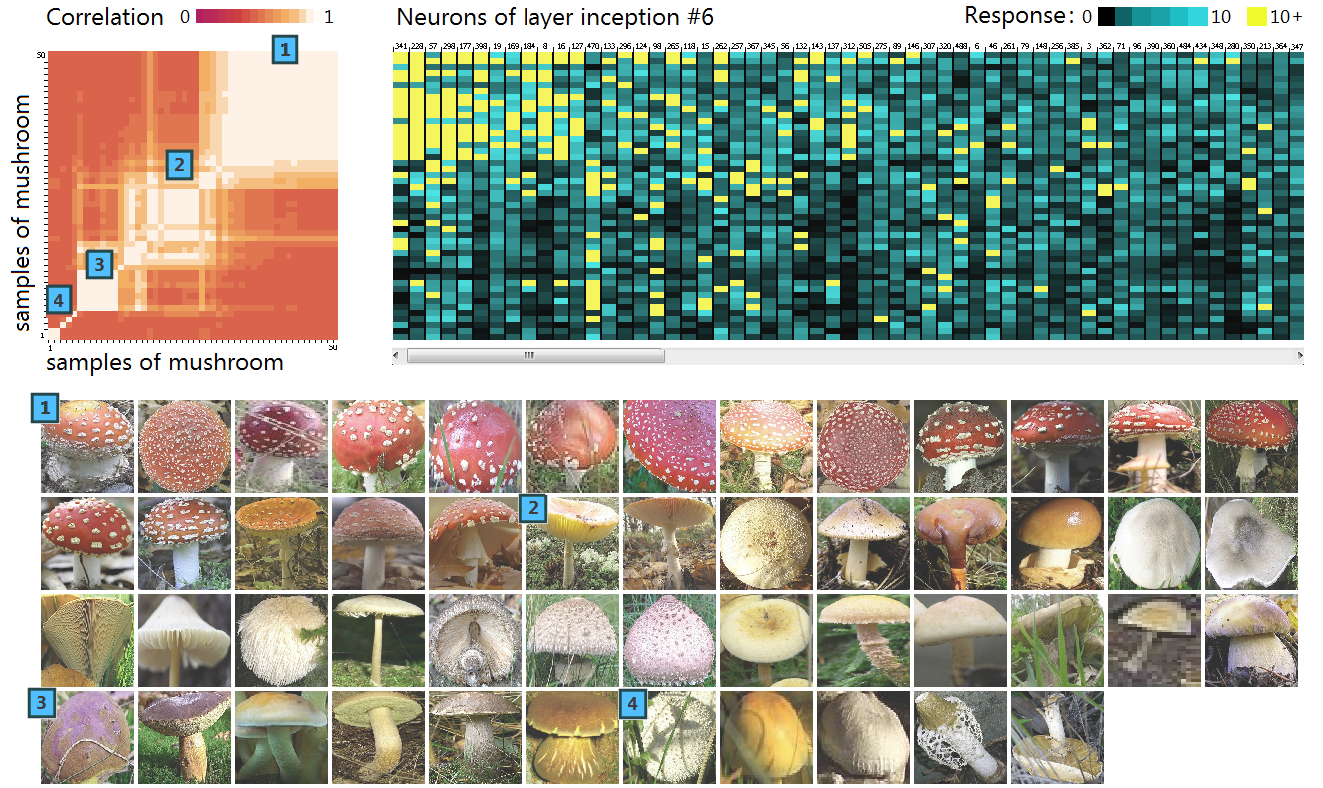}
 \caption {
	The correlation matrix between the samples of class \emph{mushroom}, along with a sample-level response map.
	Each block in the matrix corresponds to a sub-class of similar samples (e.g. red mushrooms).
 }
 \label{fig:corrmat}
\end{figure}

The response map presented above aggregates the responses per class in order to show how they vary across different classes.
In many cases, the responses vary within the same class due to latent subclasses, e.g. cut vs. full apples.
\emph{Blocks} enables exploring possible latent subclasses within the samples of a selected class in a dedicated window (Fig.~\ref{fig:corrmat}).
For this purpose, we compute the correlation matrix of network responses to these samples at a selected reference layer.
We reorder the matrix using spectral clustering and visualize it along with these responses and with thumbnails of the samples.
The responses are visualized using a sample-level response map which shows which neurons are active for which samples.
The rows in this map represent the samples, and are assigned the same order as in the correlation matrix.
The column represents the neurons of the selected reference layer.
The presence of multiple blocks in the matrix indicates the presence of latent subclasses such as different types of \emph{mushroom} (Fig.~\ref{fig:corrmat}).
Selecting a block highlights the corresponding samples and reorders the neurons according to their responses within these samples. 


By inspecting the correlation matrix at successive layers, it is possible to observe how the latent subclasses emerge in the CNN.
Despite activating different feature detectors in the CNN, these subclasses can still activate the same output unit.
This is thanks to the final layers in CNNs being fully connected, which enables the output unit of a class to combine responses from multiple features.
As noted by Nguyen et al.~\cite{nguyen2016multifaceted}, identifying latent subclasses and analyzing their properties gives opportunities to optimize the classification process (\textbf{T2}).




\subsection{Sample Viewer}
\label{sec:sampleviewer}
The sample viewer is key to inspect classification errors (\textbf{T1}) and to analyze the impact of image transformations (\textbf{T3}).
It shows thumbnail images of selected samples and offers various possibilities to manipulate and explore them (Fig.~1c).
A label at the top of the view describes what the current selection represents.
The samples can be grouped by their actual classes: a representative sample of each group is shown as thumbnail image along with a number indicating the count of the remaining samples.
This gives an overview of all classes included in the selection and helps in inferring common features among them.

When showing individual samples, the user can obtain details about them either on top of the thumbnails or in tooltips. 
For example border color can indicate whether the CNN prediction for a sample is top-1 correct, top-5 correct, or otherwise. 
%
The viewer also allows exploring saliency maps of the images to analyze the role of a selected neuron in the network.
These maps are computed using the \emph{FeatureVis} library \cite{grun2016taxonomy} and the \emph{MatConvNet} toolbox \cite{vedaldi2015matconvnet}.
They highlight image features the selected neuron responds to (Fig.~\ref{fig:featurevis}). 


The samples in the viewer can be filtered by various criteria such as membership of a selected class group, activation of a selected neuron, and class-level or group-level classification results.
Additionally, \emph{Blocks} allows loading multiple sets of classification results computed by different classifiers or after applying different data transformations.
Users can filter the samples based on these results, e.g. to show samples correctly classified under all rotations or ones correctly classified  by a selected classifier only.
This enables identifying samples and classes that have certain properties such as rotation invariance and ease of discrimination, or ones that only a selected classifier excels in.

\section{Applications}
\label{sec:apps}

The components of \emph{Blocks} offer extensive support to the analysis goals identified by Liu et al.~\cite{liu2017towards2}, as described in Section~1. 
We next demonstrate how Blocks helps in \emph{understanding} the training process, \emph{diagnosing} the separation power of the feature detectors, and \emph{improving} the architecture accordingly to yield significant gain in accuracy (\textbf{T2}).
Additionally, we illustrate how \emph{Blocks} helps in \emph{improving} the curation of training datasets by \emph{understanding} sensitivity properties of the CNN (\textbf{T3}) and \emph{diagnosing} various quality issues in the data (\textbf{T1}).


\subsection{Designing Hierarchy-Aware CNNs}
\label{sec:sepusecase}

Understanding the training behavior of CNNs helps in introducing targeted design improvements to large-class CNN classifiers.
In particular, we show how making CNNs hierarchy-aware significantly improves the accuracy and accelerates the training convergence.

\subsubsection{Understand: model convergence}

\begin{figure*}[!ht]
 \centering
 \includegraphics[width=\textwidth]{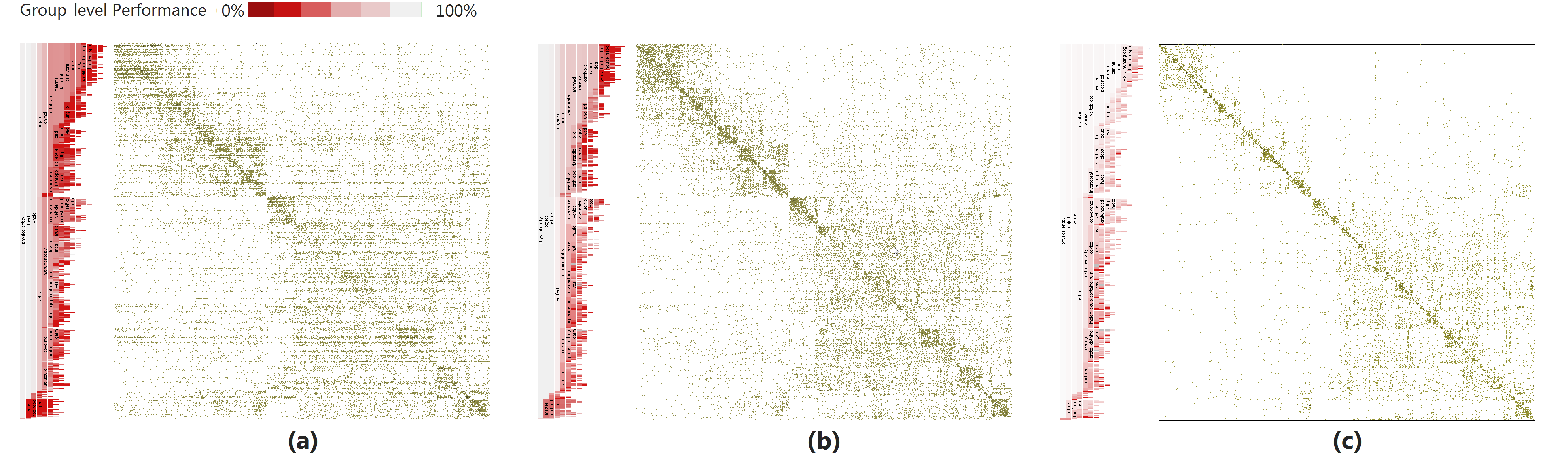}
 \caption {
	The confusion matrix after the first epoch (a), the second epoch (b), and the final epoch (c) during the training of AlexNet~\cite{krizhevsky2012imagenet}.
	The network starts to distinguish high-level groups already after the first epoch. 	The hierarchy viewers show the corresponding group-level accuracies.
 }
 \label{fig:matevolution}
\end{figure*}
The CNN classification model converges over several epochs during training phase.
We inspect the model responses at each epoch and the corresponding class confusions in the respective views in \emph{Blocks}.

Observing how the confusion matrix changes over successive epochs reveals how the final confusion patterns develop.
Initially, the model is random, resulting in a uniform distribution of the values in the confusion matrix.
Fig.~\ref{fig:matevolution}a-b depicts the confusion matrix after the first two epochs while training standard AlexNet~\cite{krizhevsky2012imagenet}.
Fig.~\ref{fig:matevolution}c depicts the matrix after the training is terminated.
It is remarkable that major blocks are already visible after only one epoch of training.
This means that the network first learns to distinguish major high-level groups such as natural objects vs. artifacts.
In the second epoch, the separation between these groups improves and subgroups within them emerge.
In the final epoch, the CNN makes fewer overall confusions that are generally limited to narrow groups.


To further analyze this behavior, we observe how the feature detectors develop during the training.
We found out that the response profiles of neurons in early layers quickly converged in the first and second epoch, with subsequent epochs leading to increasingly smaller changes.
These low-level features seem to be capable of separating high-level groups, as the confusion matrices suggest.
In contrast, the response profiles in deeper layers converged at later epochs, with changes in these epochs being increasingly limited to the last layers.
Zeiler and Fergus reported similar findings by observing the development of feature detectors during training \cite{zeiler2014visualizing}.
To confirm our observations, we next analyze the classification power of individual layers.

\subsubsection{Diagnose: feature classification power}

\emph{Blocks} allows analyzing at which layer in the CNN the feature detectors are able to separate certain groups of classes.
Each layer in the CNN abstracts the input image into a set of responses that indicate the presence of increasingly more complex features in the image.
To assess the classification power of the feature detectors at a certain layer, we train a linear classifier to classify the samples based on these features only, as proposed by Rauber et. al~\cite{rauber2017visualizing}.
This classifier characterizes each class by a weighted sum of the feature responses, and classifies a sample by computing corresponding class scores.
To analyze the performance of this linear classifier, we create a confusion matrix of its predictions.
Additionally, we color the groups in the hierarchy viewer by group-level recall.
This reveals which groups the features at each layer can already separate from each other.

We are able to confirm that the features developed at early layers can separate between high level groups with group-level performance close to the output layer.
Separating between fine-grained groups requires more sophisticated features that are developed at deeper layers.

We noticed that while AlexNet is able to separate dogs from other classes, it frequently confused certain types of dogs in ImageNet for each other (see topmost block in Fig.~1).
Szegedy et al.~\cite{szegedy2015going} argued for the need of additional convolutional layers to separate highly-similar classes.
Accordingly, their GoogLeNet CNN achieves higher accuracy than AlexNet on such classes. 
However, by comparing the performance of both CNNs, we found that GoogLeNet achieves lower accuracy for certain classes such as 'ping-pong ball' and 'horizontal bar'.
The samples of these classes are composed of simple features, which suggests that they do not benefit from deep architectures.
Moreover, we found that classifying these samples based on intermediate  features in GoogLeNet achieves higher accuracy than the output layer. 
This suggests that classification decisions should be taken at different layers in deep CNNs to account for the varying complexity of the classes.
Similar proposals were shown to improve classification accuracy such as variable-depth CNNs \cite{tan2016towards} and conditional networks \cite{ioannou2016decision}.



\subsubsection{Improve: exploiting the class hierarchy}

Our findings about model convergence and group separability at different layers enable us to improve training speed and accuracy, by involving the hierarchy information in the design and training of CNNs.

\begin{figure}[!h]
 \centering
 \includegraphics[width=\linewidth]{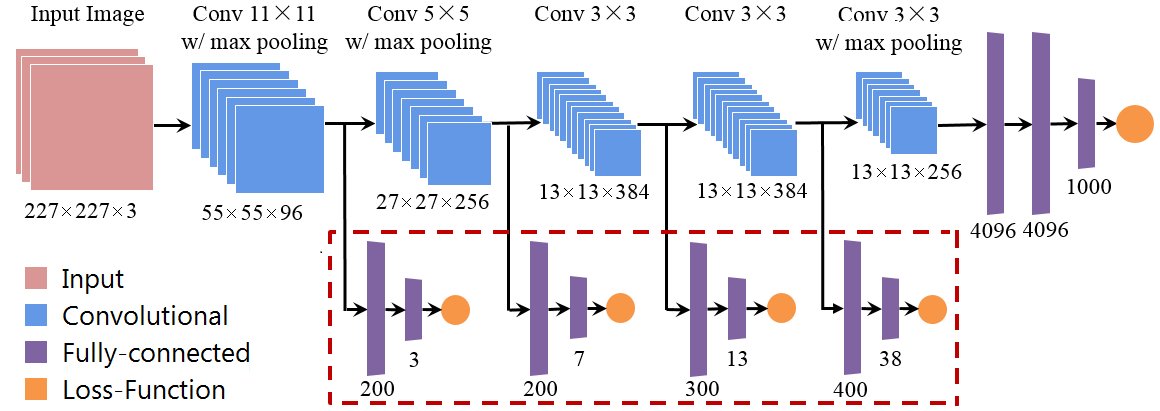}
 \caption {
	The adapted AlexNet architecture.
	The added branches are marked with a dotted box.
  These branches impose the class hierarchy during the training phase and are eliminated after training completion.
 }
 \label{fig:hierarchicalcnn}
\end{figure}

We select AlexNet~\cite{krizhevsky2012imagenet} as a reference architecture that is straigtforward to extend and re-train.
After analyzing the classification power of convolutional layers, we extended them to be hierarchy-aware.
For this purpose, we created branches from these layers that perform group-level classification and back-propagate group error (Fig.~\ref{fig:hierarchicalcnn}).
We require the first layer to classify the samples into 3 broad groups only, and increased the number of groups in subsequent layers.
For each layer, we selected groups that we identified as most separable using the corresponding feature detectors.
These groups, along with the trained model are provided in the supplementary material.

We re-train the adapted network on the ILSVRC dataset for 50 epochs using Caffe \cite{jia2014caffe}.
Table~\ref{table:perf} summarizes the validation error at epoch 25, compared with baseline AlexNet. The results did not improve beyond this epoch.
\begin{table}[h]
\begin{center}
\def\arraystretch{1.15}
  \begin{tabular}{  r | c c}
    Architecture						&	Top-1 error	& Top-5 error	\\ \hline
    Standard AlexNet				& $42.6\%$		& $19.6\%$		\\
    Hierarchy-Aware AlexNet	& $\textbf{34.33}\%$			& $\textbf{13.02}\%$		\\ 
  \end{tabular}
\caption{Performance of baseline vs. improved architectures.}
\label{table:perf}
\end{center}
\end{table}

The hierarchy-aware architecture cuts the top-5 error down by more than one third.
The classification results are computed from the main branch of the network, which is identical in complexity to baseline AlexNet.
The additional branches play no role after the training is completed.
This means that the improved accuracy can be fully attributed to involving the hierarchy information during training.

Our results show more significant improvement on the ILSVRC dataset than \emph{HD-CNN}, a recently proposed approach to implement hierarchical CNNs \cite{yan2015hd}.
This shows the value of understanding the separation power of each layer and of 
introducing the hierarchy information accordingly.
This is especially beneficial when the network is deep and the number of classes is large. 
Furthermore, the model converged quickly in our experiment, with top-5 error reaching $24.6\%$ only after 4 epochs.
This is because the additional loss functions directly update the weights of
the corresponding layers to achieve group separation.
This offers new solutions to the vanishing gradient problem in deep models \cite{lecun2012efficient}.
Moreover, this aids generalizability since our trained model should satisfy multiple loss functions and is hence less likely to overfit the training data than standard CNNs.




\subsection{Sensitivity to Image Transformations} 
\label{sec:sensitivity}

The classes in ImageNet vary in their sensitivity to image transformations.
In the following we analyze the impact of gray-scale color conversion and image rotation on classification accuracy.
This reveals whether the corresponding features are invariant to color and rotation.


\subsubsection{Color invariance}

\begin{figure}[!b]
 \centering
 \includegraphics[width=\linewidth]{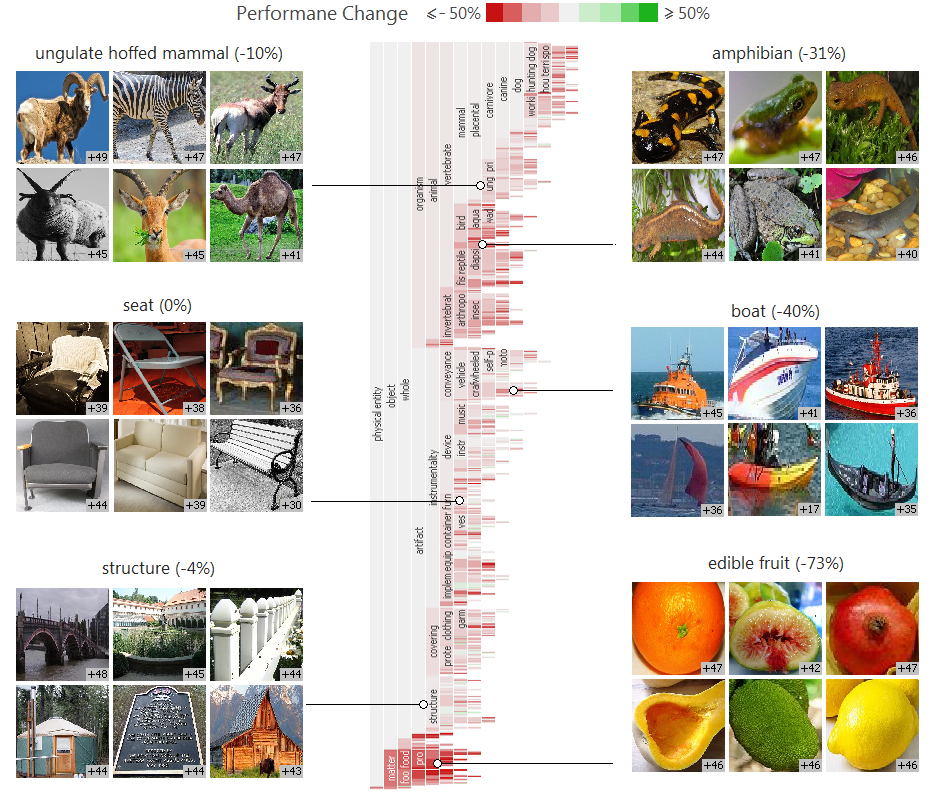}
 \caption {
	Color-invariant (left) vs. color-sensitive classes (right). 
 }
 \label{fig:colorinv}
\end{figure}

We convert the images in the ILSVRC validation dataset into grayscale and re-classify them using GoogLeNet.
Figure~\ref{fig:colorinv} shows the impact of this transformation on the classification results.
The hierarchy viewer depicts change in group-level precision for each group in the hierarchy, compared with the result of original color images.
Red indicates a drop in the accuracy due to the lack of color information.

The largest absolute drop can be observed in the \emph{food} groups such as \emph{fruits} ($-60\%$), \emph{vegetables} ($-43\%$), and \emph{dishes} ($-67\%$).
By inspecting the confusion matrix, we found out that the CNN confuses these samples mainly for classes in other groups such as \emph{tableware}, \emph{cookware}, \emph{covering}, \emph{containers}, \emph{fungus}, and \emph{fishes}.
In contrast, most artifact groups and classes had minimal or no change in accuracy such as \emph{electronic equipment} ($0\%$), \emph{seats} ($0\%$), \emph{measuring instruments} ($-1\%$), \emph{wheeled vehicles} ($-3\%$) and \emph{structures} ($-3\%$).
By inspecting the training samples in these groups, we found strong variation in color. 
This enforces the CNN to rely on color-independent features to recognize these classes.
Some exceptions were lifeboats ($-84\%$), tennis balls ($-58\%$), jack-o'-laterns ($-48\%$), and lipsticks ($-42\%$), all of which had training samples of particular color.
\begin{figure}[!th]
 \centering
 \includegraphics[width=\linewidth]{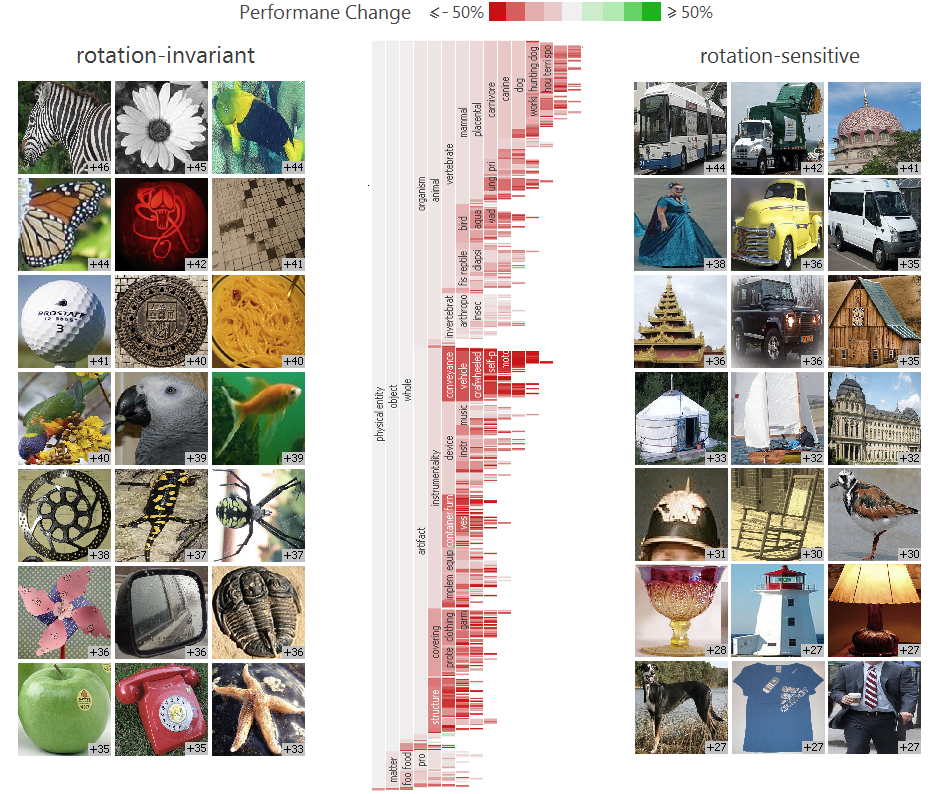}
 \caption {
	Rotation-invariant (left) vs. rotation-sensitive classes (right). 
 }
 \label{fig:rotinv}
\end{figure}
By inspecting the corresponding features we found that the CNN relies on color-dependent features as discriminative common denominators of the corresponding samples, even if these samples have distinctive shapes.

After inspecting the changes in accuracy, the curators of training data can alleviate color dependence by including grayscale versions or additional samples of the impacted classes to balance color variation.
Alternatively, the CNN architecture can be adapted to simulate rods and cones in natural vision.
Color information remains necessary, however, to recognize classes with intrinsic color that are otherwise hard to distinguish from similar classes such as \emph{green snakes}.

\subsubsection{Rotation invariance}

We re-classify the images in ILSVRC after rotating them by $90^{\circ}$ and observe the change in group-level accuracy as in the previous section.
By observing rotation-invariant classes (Fig.~\ref{fig:rotinv}-left), we found that they often have circular shapes as with \emph{ball} and \emph{flower}, or have rotation-invariant features based on texture and color as with \emph{zebra} and various \emph{produce} classes.
On the other hand, rotation-sensitive classes (Fig.~\ref{fig:rotinv}-right) have non-circular shapes and mostly appear in a specific pose as with the \emph{vehicles} and \emph{buildings}.
Accordingly the latter groups exhibit significant drop in accuracy of $-55\%$ and $-50\%$ respectively.

Among animals \emph{invertebrates} exhibit the lowest drop of $4\%$, although they do not have the circular shape.
By inspecting the corresponding training samples, we found that the objects exist in various rotations, which led the CNN to develop rotation-invariant features as common denominators among the samples of each class.
In contrast, most samples of \emph{aquatic birds} ($-39\%$) and \emph{hoofed mammals} ($-45\%$) did have the same pose, leading to rotation-sensitive features.





\subsection{Inspecting Data Quality}

\label{sec:data_quality}
The classes in the ILSVRC dataset were chosen randomly from the WordNet ontology.
Filtering the confusion matrix to show frequent confusions and inspecting the remaining block outliers reveals several issues with the choice of these classes (Fig.~\ref{fig:matrix_filtering}), such as:
\begin{itemize}
	\item Redundancy:
		  two classes are identical but belong to different WordNet branches
			such as \emph{missile} and \emph{projectile, missile}, \emph{bassinet} and \emph{cradle}, or \emph{sunglass} and \emph{sunglasses, dark glasses}.
	\item Subclass relations: one class is a special type of the other class such as 
		\emph{bolete} and \emph{mushroom}, or \emph{coffee mug} and \emph{cup}.
	\item Part-of relationships: one class represents part of another class such as
	\emph{wing} and \emph{airplane}, or \emph{monitor} and \emph{desktop computer}.
	\item Semantic ambiguity: two classes have similar semantics such as 
	\emph{bookstore} and \emph{library}, or \emph{gas mask} and \emph{oxygen mask}.
	\item Abstract classes: one class such as \emph{groom} takes multiple forms that
	are often confused with physical classes such as \emph{suit}.
\end{itemize}
These issues impact about $5\%$ of the classes, and lead to a significant drop in the top-1 classification accuracy which is not caused by the classifier.
Nevertheless, they apparently remained largely unnoticed due to reliance on top-5 error to compare classifiers.
This error measure, however, intends to account for images that actually contain multiple objects, and is usually not used during the training phase.
Ensuring non-overlapping class semantics helps in sharpening their feature detectors and improving the overall performance accordingly.

\emph{Blocks} also helps in detecting mislabeled samples such as an image of a lion labeled as monkey.
We found such cases by inspecting misclassified samples having very high prediction probability and very low probability assigned to the ground truth.
Isolating such cases is useful to robustly compare different architectures.
Finally, \emph{Blocks} helps in restructuring the pre-defined class hierarchy of ImageNet to better reflect their visual similarity structures.
For example, the groups \emph{fruit} and \emph{editable fruit} belong to different branches of the hierarchy root despite having high similarity, which led to  frequent inter-group confusions.



\section{Discussion}


\label{sec:discussion}

\emph{Blocks} is the first system to enable analyzing the impact of a class hierarchy on CNNs and improving their design accordingly.
Next we discuss how \emph{Blocks} relates to previous work, summarize its limitations, and report feedback of deep-learning experts on our system.

\subsection{Related Work}

Previous work has utilized similar visualizations to the ones in \emph{Blocks}, focusing, however, on different data facets or tasks.

\paragraph{Confusion matrices} have been utilized to manipulate decision boundaries as in \emph{ManiMatrix} \cite{kapoor2010interactive}, to combine multiple classifiers as in \emph{EnsembleMatrix} \cite{talbot2009ensemblematrix}, and to examine impact of model changes as in BaobabView~\cite{van2011baobabview}.
Little focus has been given to revealing nested block patterns in these matrices, unlike matrices showing correlations \cite{wang2002classification} or distances \cite{brasselet2009optimal} between the samples.
Alternatives to confusion matrices have focused on prediction probabilities \cite{alsallakh2014visual, amershi2015modeltracker, cao2016untangle, ren2017squares} or on the ground truth \cite{beauxis2014visualization}, and hence do not involve the class hierarchy.

\paragraph{Heatmaps} have also been used to visualize selected responses for single samples both in the input space \cite{jaderberg2014deep} and in the class space \cite{bendale2016towards}.
\emph{CNNVis} utilizes a \emph{class}~$\times$~\emph{neuron} response map to show activation patterns within certain neuron groups \cite{liu2017towards}.
Nevertheless, these maps are not designed to provide a comprehensive overview of the responses or to reveal group-level response patterns, a key focus of \emph{Blocks}.

\paragraph{Sample viewers} are often integrated in machine learning environments to inspect individual samples along with their attributes and models responses \cite{alsallakh2014visual, brooks2015featureinsight, patel2010gestalt}.
\emph{LSTMVis}~\cite{strobelt2016visual} features a powerful viewer for text data.
It allows comparing multiple sentences to reveal linguistic properties captured by each hidden state in LSTM neural networks.
In contrast, available viewers for CNN-based image classification data have focused mainly on visualizing image features for a few samples \cite{bruckner2014ml, liu2017towards, yosinski2015understanding}.
Unlike \emph{Blocks}, they lack possibilities to explore a large number of samples and compare multiple result sets.

\subsection{Scalability and Limitations}

As we demonstrated in previous sections, \emph{Blocks} supports analyzing complex CNN classifiers such as AlexNet~\cite{krizhevsky2012imagenet} and GoogLeNet~\cite{szegedy2015going}, trained to classify datasets at the scale of ILSVRC (Section~\ref{sec:dataset}).


The grouping of the classes is vital to support scalability with the number of classes.
High-level groups are easy to identify in the hierarchy viewer, as their labels can be depicted.
Small groups can still be identified interactively by means of tooltips.
Selecting one of these groups shows thumbnails of its classes in the sample viewer, which in turn makes these classes easy to identify and select individually.

The confusion matrix view can handle a $1000 \times 1000$ matrix without need for scrolling.
Multiscale aggregation \cite{elmqvist2008zame} enables handling larger matrices, thanks to similarity-based class ordering.
While this does not show confusion between individual classes, it provides overview of major block patterns and block outliers. 

The response map can provide overview of neuron responses in a selected layer to a large number of samples, thanks to per-class aggregation and downsampling.
A typical intermediate layer in the CNNs we examined contains about 512 $4 \times 4$ filters.
A standard $1920 \times 1080$ display can hence fit about $15-20\%$ of the corresponding response profiles, along with the auxiliary views.
This is sufficient to explore the most relevant profiles for selected classes or groups, thanks to relevance-based ordering.

Besides scalability limits, \emph{Blocks} is also limited in the data facets it shows in CNNs.
Unlike \emph{CNNVis}, \emph{Blocks} does not provide information about layer connectivity and hence does not reveal patterns in the connection weights.
Furthermore, the layer responses are visualized independently for each layer.
This hinders close inspection of how the CNN develops the feature detectors, in particular how the detectors in one layer rely on the ones in previous layers.
We envision that combining features from \emph{Block} and \emph{CNNVis} might provide such possibilities.

Finally, \emph{Blocks} currently offers few possibilities to monitor the training process, limited to changes in the confusion matrix and response map.
Further work is needed to closely examine the impact of various training parameters on the CNN features and performance, including initialization strategies such as pre-training \cite{erhan2010does, erhan2009difficulty}, learning rate, and regularization strategies such as DropOut \cite{srivastava2014dropout}.


Except for the response map, the views in \emph{Blocks} are not restricted to CNN classifiers.
Furthermore, this map can visualize internal responses of any classifier that is based on a number of feature detectors.
This makes \emph{Blocks} a potentially generic tool to analyze large-class classifiers, focusing on how an explicit or latent class hierarchy impacts the classification model and performance.

\subsection{Expert Feedback}
We solicited feedback on our system from an external expert in CNNs who developed various CNN visualization systems \cite{nguyen2016synthesizing, nguyen2015deep, nguyen2016multifaceted, yosinski2015understanding}.
He finds `\emph{the visualizations are easy to follow and make sense}' and `\emph{the idea of comparing the classes along the hierarchy is novel}'.
He further comments: `\emph{I have not seen a tool that puts all these really useful features together!
Traditionally, one would have to write code to perform these analyses manually. This tool would be incredibly useful and advance science further.}'
These insights we report in Section~\ref{sec:apps} demonstrate the value of the visualization, as proposed by Stasko \cite{stasko2014value}.
A further study is needed to assess the usability of our system.


\section{Conclusion and Future Work}
\label{sec:conclusion}

We presented visual-analytics methods to inspect CNNs and to improve their design and accuracy on large-scale image classification.
Our methods are based on identifying the hierarchical similarity structures between the classes as key information that impacts various properties of CNNs.
These structures influence the feature detectors developed by the CNN at different layers and over different training epochs.
We demonstrated how understanding these influences help in designing hierarchy-aware CNN architectures that yield significant gain in classification accuracy and in convergence speed.
We further demonstrate how extracting and analyzing the class similarity structure can reveal various quality issues in the training dataset such as overlapping class semantics, labeling issues, and imbalanced distributions.
This is key to improve the CNN robustness to data variation by curating a representative dataset.
Our future work aims to study how class similarity structures influence other types of large-scale classifiers and how our findings can be generalized to domains other than image classification.



\acknowledgments{
 We thank Jitandra Malik for encouraging us to pursue our initial ideas, Anh Nguyen for feedback and Felix Gr{\"u}n for help on FeatureVis. 
}

\bibliographystyle{abbrv}
\bibliography{main}
\end{document}